\documentclass{article}
\PassOptionsToPackage{numbers, compress}{natbib}

\usepackage[final]{neurips_2024}

\usepackage[utf8]{inputenc} 
\usepackage[T1]{fontenc}    
\usepackage{hyperref}       
\usepackage{url}            
\usepackage{booktabs}       
\usepackage{amsfonts}       
\usepackage{nicefrac}       
\usepackage{microtype}      
\usepackage{xcolor}         
\usepackage{graphicx}
\usepackage{amsmath}
\usepackage{algorithm}
\usepackage{algpseudocode}
\usepackage{xparse}
\usepackage{footmisc}
\usepackage{float}

\title{BatchTopK Sparse Autoencoders}

\author{%
    Bart Bussmann \\
    Independent \\
    \texttt{bartbussmann@gmail.com} \\
            \And
    Patrick Leask \\
    Department of Computer Science \\
    Durham University \\
    \texttt{patrick.leask@durham.ac.uk} \\
    \And 
    Neel Nanda
}

\begin{document}

\maketitle


\begin{abstract}
Sparse autoencoders (SAEs) have emerged as a powerful tool for interpreting language model activations by decomposing them into sparse, interpretable features. A popular approach is the TopK SAE, that uses a fixed number of the most active latents per sample to reconstruct the model activations. We introduce BatchTopK SAEs, a training method that improves upon TopK SAEs by relaxing the top-k constraint to the batch-level, allowing for a variable number of latents to be active per sample. As a result, BatchTopK adaptively allocates more or fewer latents depending on the sample, improving reconstruction without sacrificing average sparsity. We show that BatchTopK SAEs consistently outperform TopK SAEs in reconstructing activations from GPT-2 Small and Gemma 2 2B, and achieve comparable performance to state-of-the-art JumpReLU SAEs. However, an advantage of BatchTopK is that the average number of latents can be directly specified, rather than approximately tuned through a costly hyperparameter sweep.  We provide code for training and evaluating BatchTopK SAEs at \url{https://github.com/bartbussmann/BatchTopK}.


\end{abstract}

\section{Introduction}

Sparse autoencoders (SAEs) have been proven effective for finding interpretable directions in the activation space of language models \cite{bricken2023towards, cunningham2023sparse, templeton2024scaling, rajamanoharan2024improvingdictionarylearninggated}. SAEs find approximate, sparse, linear decompositions of language model activations by learning a dictionary of interpretable latents from which the activations are reconstructed. 

The objective used in training SAEs \cite{bricken2023towards} has both a sparsity and a reconstruction term. These are naturally in tension as, for an optimal dictionary of a given size, improving the reconstruction performance requires decreasing sparsity and vice versa. Recently, new architectures have been proposed to address this issue, and achieve better reconstruction performance at a given sparsity level, such as Gated SAEs \cite{rajamanoharan2024improvingdictionarylearninggated}, JumpReLU SAEs \cite{rajamanoharan2024jumpingaheadimprovingreconstruction}, and TopK SAEs \cite{gao2024scalingevaluatingsparseautoencoders}. 

In this paper, we introduce BatchTopK SAEs, a novel variant that extends TopK SAEs by relaxing the top-$k$ constraint to a batch-level constraint. This modification allows the SAE to represent each sample with a variable number of latents, rather than assuming that all model activations consist of the same number of units of analysis. By selecting the top activations across the entire batch, BatchTopK SAEs enable more flexible and efficient use of the latent dictionary, leading to improved reconstruction without sacrificing average sparsity. During inference we remove the batch dependency by estimating a single global threshold parameter.

Through experiments on the residual streams of GPT-2 Small \cite{radford2019language} and Gemma 2 2B \cite{team2024gemma}, we show that BatchTopK SAEs consistently outperform both TopK and JumpReLU SAEs in terms of reconstruction performance across various dictionary sizes and sparsity levels, although JumpReLU SAEs have less downstream CE degradation in large models with a high number of active latents. Moreover, unlike JumpReLU SAEs, BatchTopK SAEs allow direct specification of the sparsity level without the need for tuning additional hyperparameters.

\section{Background: Sparse Autoencoder Architectures}

\begin{figure}
    \centering
    \includegraphics[width=0.9\linewidth]{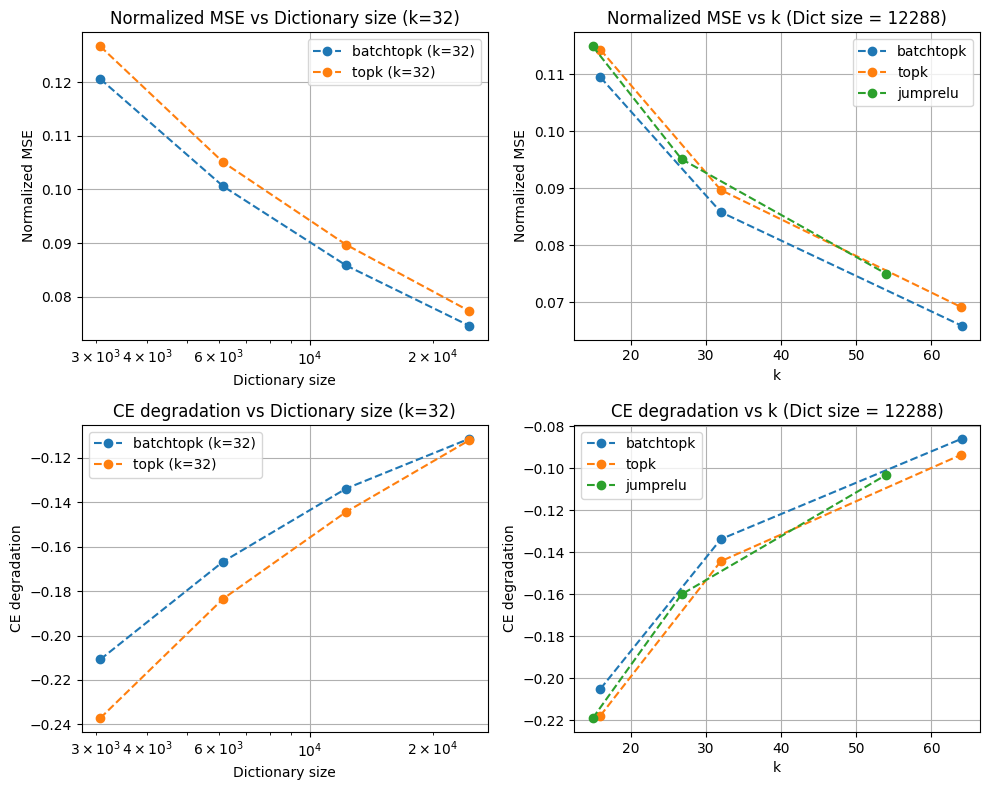}
    \caption{On GPT-2 Small activations, BatchTopK largely achieves better NMSE and CE than standard TopK across different dictionary sizes, for a fixed number of active latents of 32 (Left). JumpReLU SAEs are omitted from this comparison as their L0 cannot be fixed to a value. For fixed dictionary size (12288) and varying levels of k, BatchTopK outperforms TopK and JumpReLU SAES, both in terms of NMSE and CE (Right).}
    \label{fig:batchtopkvstopk}
\end{figure}

Sparse autoencoders aim to learn efficient representation of data by reconstruction inputs while enforcing sparsity in the latent space. In the context of large language models, SAEs decompose model activations $\mathbf{x} \in \mathbb{R}^n$ into sparse linear combinations of learned directions, which are often interpretable and monosemantic. 

An SAE consists of an encoder and a decoder:

\begin{align}
    \mathbf{f}(\mathbf{x}) &:= \sigma(\mathbf{W}_\text{enc}\mathbf{x} + \mathbf{b}_\text{enc}), \label{eq:1} \\[0.5em]
    \hat{\mathbf{x}}(\mathbf{f(x)}) &:= \mathbf{f(x)} \mathbf{W}_\text{dec} + \mathbf{b}_\text{dec}. \label{eq:2}
\end{align}

where $\mathbf{f}(\mathbf{x}) \in \mathbb{R}^m$ is the sparse latent representation and $\hat{\mathbf{x}}(\mathbf{f}) \in \mathbb{R}^n$ is the reconstructed input. $\mathbf{W}^\text{enc}$ is the encoder matrix with dimension $n \times m$ and  $\mathbf{b}^\text{enc}$ is a vector of dimension $m$; conversely $\mathbf{W}^\text{dec}$ is the decoder matrix with dimension $m \times n$ and  $\mathbf{b}^\text{dec}$ is of dimension $n$.  The activation function $\sigma$ enforces non-negativity and sparsity in $\mathbf{f}(\mathbf{x})$, and a latent $i$ is active on a sample $\mathbf{x}$ if $f_i({\mathbf{x}}) > 0$.

SAEs are trained on the activations of a language model at a particular site, such as the residual stream, on a large text corpus, using a loss function of the form 

\begin{equation}
    \mathcal{L}(\mathbf{x}) := \underbrace{\|\mathbf{x} - \hat{\mathbf{x}}(\mathbf{f}(\mathbf{x}))\|_2^2}_{\mathcal{L}_\text{reconstruct}} + \underbrace{\lambda \mathcal{S}(\mathbf{f}(\mathbf{x}))}_{\mathcal{L}_\text{sparsity}} + \alpha \mathcal{L}_\text{aux}
\end{equation}

where $\mathcal{S}$ is a function of the latent coefficients that penalizes non-sparse decompositions, and $\lambda$ is a sparsity coefficient, where higher values of $\lambda$ encourage sparsity at the cost of higher reconstruction error. Some architectures also require the use of an auxiliary loss $\mathcal{L}_\text{aux}$, for example to recycle inactive latents in TopK SAEs.

\textbf{ReLU SAEs} \cite{bricken2023towards} use the L1-norm $S(\boldsymbol{f}) := || \boldsymbol{f} ||_1$ as an approximation to the L0-norm for the sparsity penalty. This provides a gradient for training unlike the L0-norm, but suppresses latent activations harming reconstruction performance \cite{rajamanoharan2024improvingdictionarylearninggated}. Furthermore, the L1 penalty can be arbitrarily reduced through reparameterization by scaling the decoder parameters, which is resolved in \cite{bricken2023towards} by constraining the decoder directions to the unit norm. Resolving this tension between activation sparsity and value is the motivation behind the newer architecture variants.

\textbf{TopK SAEs} \cite{gao2024scalingevaluatingsparseautoencoders, makhzani2014ksparseautoencoders} enforce sparsity by retaining only the top $k$ activations per sample. The encoder is defined as:

\begin{equation}
    \boldsymbol{f}(\boldsymbol{x}) := \text{TopK}(\mathbf{W}_\text{enc}\mathbf{x} + \mathbf{b}_\text{enc})
\end{equation}

where $\text{TopK}$ zeroes out all but the $k$ largest activations in each sample. This approach eliminates the need for an explicit sparsity penalty but imposes a rigid constraint on the number of active latents per sample. An auxiliary loss $\mathcal{L}_{aux} = || e - \hat{e} ||^2$ is used to avoid dead latents, where $\hat{e} = W_{dec} z$ is the reconstruction using only the top-$k_{aux}$ dead latents (usually 512), this loss is scaled by a small coefficient $\alpha$ (usually 1/32). 

\textbf{JumpReLU SAEs} \cite{rajamanoharan2024jumpingaheadimprovingreconstruction} replace the standard ReLU activation function with the JumpReLU activation, defined as

\begin{equation}
    \text{JumpReLU}_\theta(z) := zH(z-\theta)
\end{equation}

where $H$ is the Heaviside step function, and $\theta$ is a learned parameter for each SAE latent, below which the activation is set to zero. JumpReLU SAEs are trained using a loss function that combines L2 reconstruction error with an L0 sparsity penalty, using straight-through estimators to train despite the discontinuous activation function. A major drawback of the sparsity penalty used in JumpReLU SAEs compared to (Batch)TopK SAEs is that it is not possible to set an explicit sparsity and targeting a specific sparsity involves costly hyperparameter tuning. While evaluating JumpReLU SAEs,  \cite{rajamanoharan2024jumpingaheadimprovingreconstruction} chose the SAEs from their sweep that were closest to the desired sparsity level, however this resulted in SAEs with significantly different sparsity levels being directly compared. JumpReLU SAEs use no auxiliary loss function.

\section{BatchTopK Sparse Autoencoders}

We introduce \textbf{BatchTopK SAEs} as an improvement over standard TopK SAEs. In BatchTopK, we replace the sample-level $\text{TopK}$ operation with a batch-level $\text{BatchTopK}$ function. Instead of selecting the top $k$ activations for each individual sample, we select the top $n \times k$ activations across the entire batch of $n$ samples, setting all other activations to zero. This allows for a more flexible allocation of active latents, where some samples may use more than $k$ latents while others use fewer, potentially leading to better reconstructions of the activations that are more faithful to the model.

The training objective for BatchTopK SAEs is defined as:

\begin{equation}
\mathcal{L}(\mathbf{X}) = \|\mathbf{X} - \text{BatchTopK}(\mathbf{W}_{enc}\mathbf{X} + \mathbf{b}_{enc})\mathbf{W}_{dec} + \mathbf{b}_{dec})\|_2^2 + \alpha \mathcal{L}_\text{aux}
\end{equation}

Here, $\mathbf{X}$ is the input data batch; $\mathbf{W}_{\text{enc}}$ and $\mathbf{b}_{\text{enc}}$ are the encoder weights and biases, respectively; $\mathbf{W}_{\text{dec}}$ and $\mathbf{b}_{\text{dec}}$ are the decoder weights and biases. The BatchTopK function sets all activation values to zero that are not among the top $n \times k$ activations by value in the batch, not changing the other values. The term $\mathcal{L}_\text{aux}$ is an auxiliary loss scaled by the coefficient $\alpha$, used to prevent dead latents, and is the same as in TopK SAEs.

BatchTopK introduces a dependency between the activations for the samples in a batch. We alleviate this during inference by using a threshold $\theta$ that is estimated as the average of the minimum positive activation values across a number of batches:

\begin{equation}
\theta = \mathbb{E}_\mathbf{X} [\min \{z_{i,j}(\mathbf{X}) | z_{i,j}(\mathbf{X}) > 0 \}]
\end{equation}

where $z_{i,j}(\mathbf{X})$ is the $j$th latent activation of the $i$th sample in a batch $\mathbf{X}$. With this threshold, we use the JumpReLU activation function during inference instead of the BatchTopK activation function, zeroing out all activations under the threshold $\theta$.

 \begin{figure}
    \centering
    \includegraphics[width=1.\linewidth]{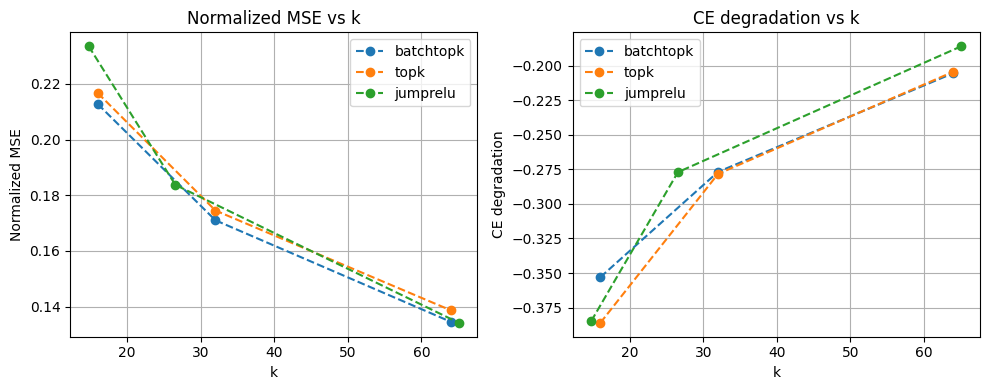}
    \caption{On Gemma 2 2B activations, BatchTopK outperforms TopK SAEs across different values of k. Although BatchTopK has a better reconstruction performance (left), it only outperforms JumpReLU in terms of downstream CE degradation in the setting where k=16 (right).}
    \label{fig:gemma_comparison}
\end{figure}

\section{Experiments}
We evaluate the performance of BatchTopK on the activations of two LLMs: GPT-2 Small (residual stream layer 8) and Gemma 2 2B (residual stream layer 12). We use a range of dictionary sizes and values for $k$, and compare our results to TopK and JumpReLU SAEs in terms of normalized mean squared error (NMSE)  and cross-entropy (CE) degradation. For the experimental details, see Appendix \ref{app:details}.

We find that for a fixed number of active latents (L0=32) the BatchTopK SAE has a lower normalized MSE and less cross-entropy degradation than TopK SAEs on both GPT-2 activations (Figure \ref{fig:batchtopkvstopk}) and Gemma 2 2B (Figure \ref{fig:gemma_comparison}). Furthermore, we find that for a fixed dictionary size (12288) BatchTopK outperforms TopK for different values of k on both models.

In addition, BatchTopK outperforms JumpReLU SAEs on both measures on GPT-2 Small model activations at all levels of sparsity. On Gemma 2 2B model activations the results are more mixed: although BatchTopK achieves better reconstruction than JumpReLU for all values of k, BatchTopK only outperforms JumpReLU in terms of CE degradation in the lowest sparsity setting (k=16).

\begin{figure}[h!]
    \centering
    \includegraphics[width=0.75\linewidth]{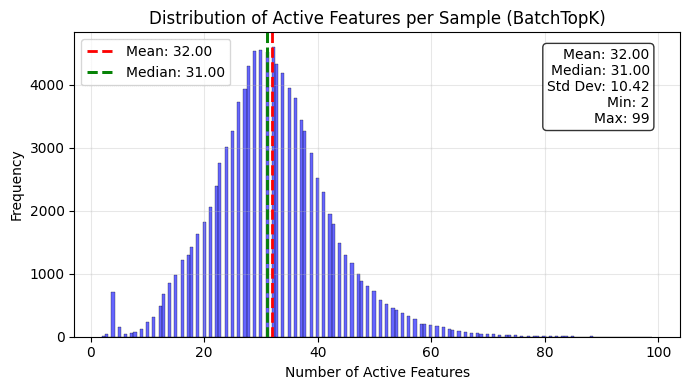}
    \caption{Distribution of the number of active latents per sample for a BatchTopK model. The distribution shows that some samples use very few latents, while others use many, illustrating the flexibility that BatchTopK provides.}
    \label{fig:activefeaturespersample}
\end{figure}

To confirm that BatchTopK SAEs make use of the enabled flexibility to activate a variable number of latents per sample, we plot the distribution of the number of active latents per sample in Figure \ref{fig:activefeaturespersample}. We observe that BatchTopK indeed uses a wide range of active latents, activating only a single latent on some samples and activating more than 80 on others. The peak on the left of the distribution are model activations on the <BOS>-token. This serves as an example of the advantage of BatchTopK: when the model activations do not contain much information, BatchTopK does not activate many latents, whereas TopK would use the same number of latents regardless of the input. This corroborates our hypothesis that the fixed TopK in \cite{gao2024scalingevaluatingsparseautoencoders} is too restrictive and that samples contain a variable number of active dictionary latents.

\section{Conclusion}

In this work, we introduced BatchTopK sparse autoencoders, a novel variant of TopK SAEs that relaxes the fixed per-sample sparsity constraint to a batch-level constraint. By selecting the top activations across the entire batch rather than enforcing a strict limit per sample, BatchTopK allows for a variable number of active latents per sample. This flexibility enables the model to allocate more latents to complex samples and fewer to simpler ones, thereby improving overall reconstruction performance without sacrificing average sparsity. We evaluated BatchTopK SAEs using the standard metrics of reconstruction loss and sparsity. We evaluated BatchTopK SAEs using standard metrics of reconstruction loss and sparsity, and while we did not directly assess human interpretability, the architectural similarity to TopK SAEs suggests that these latents would remain comparably interpretable. Our results demonstrate that small modifications to the activation function can have significant effects on SAE performance and expect that future work will continue to find improvements that better approximate the latent structure of model activations.

\section*{Acknowledgements}
We thank the ML Alignment and Theory Scholars (MATS) program for facilitating this work. In particular, we are grateful for McKenna Fitzgerald for her support in planning and managing this collaboration. We are also thankful for the helpful discussions we had with Joseph Bloom, Curt Tigges, Adam Karvonen, Can Rager, Javier Ferrando, Oscar Obeso, Stepan Shabalin, and Slava Chalnev about this work. Finally, BB and PL want to thank AI Safety Support for funding this work.

\bibliography{refs}

\appendix \section{Supplemental Material}

\subsection{Experimental Details}\label{app:details}

In this appendix, we provide details about the datasets used, model configurations, and hyperparameters for our experiments.

We trained our sparse autoencoders (SAEs) on the OpenWebText dataset\footnote{\url{https://huggingface.co/datasets/openwebtext}}, which was processed into sequences of a maximum of 128 tokens for input into the language models.

All models were trained using the Adam optimizer with a learning rate of $3 \times 10^{-4}$, $\beta_1 = 0.9$, and $\beta_2 = 0.99$. The batch size was 4096, and training continued until a total of $1 \times 10^9$ tokens were processed. 

We experimented with dictionary sizes of 3072, 6144, 12288, and 24576 for the GPT-2 Small model, and used a dictionary size of 16384 for the experiment on Gemma 2 2B. In both experiments, we varied the number of active latents $k$ among 16, 32, and 64. 

For the JumpReLU SAEs, we varied the sparsity coefficient such that the resulting sparsity would match the active latents $k$ of the BatchTopK and TopK models. The sparsity penalties in the experiments on GPT-2 Small were 0.004, 0.0018, and 0.0008. For the Gemma 2 2B model we used sparsity penalties of 0.02, 0.005, and 0.001. In both experiments, we set the bandwidth parameter to 0.001.

\end{document}